\documentclass[11pt]{article}

\usepackage[margin=1in]{geometry}
\usepackage{times}
\usepackage{amsmath,amssymb}
\usepackage{graphicx}
\usepackage{booktabs}
\usepackage[numbers,sort&compress]{natbib}
\usepackage{hyperref}
\usepackage{caption}
\usepackage{xcolor}
\usepackage[most]{tcolorbox}

\hypersetup{colorlinks=true, linkcolor=blue, citecolor=blue, urlcolor=blue}

\title{Compress and Forget: bitsandbytes Quantization Amplifies Proactive Interference in LLMs}

\author{
\begin{minipage}[t]{0.45\textwidth}
\centering
Shayan Shahrabi-Farahani \\[2pt]
Shahid Beheshti University \\[2pt]
\texttt{sh.shahrabifarahani@mail.sbu.ac.ir}
\end{minipage}
\hfill
\begin{minipage}[t]{0.45\textwidth}
\centering
Dara Rahmati \\[2pt]
Shahid Beheshti University, Tehran, Iran \\[2pt]
\texttt{d\_rahmati@sbu.ac.ir}
\end{minipage}
}

\date{}

\begin{document}
\maketitle

\begin{abstract}
Proactive Interference (PI) is a documented LLM failure mode in which retrieval of a repeatedly overwritten value degrades 
as prior overwrites accumulate, mirroring a classical phenomenon in human working memory. Post-training quantization (PTQ) is now the default deployment path for open-weight models, yet its effect on this failure 
mode has not been tested. We evaluate three precision levels (FP16, INT8, INT4/NF4, via bitsandbytes) across three 
architecturally distinct instruction-tuned models (Qwen2.5-7B-Instruct, Mistral-7B-Instruct-v0.3, Phi-3.5-mini-instruct), 
holding the retrieval task fixed. It turns out that INT4 quantization significantly reduces accuracy under high interference 
in every model (e.g., 81.0\% $\to$ 68.3\% for Qwen), confirmed by paired McNemar's tests ($p \le 2.6\times10^{-6}$) 
and a mixed-effects regression spanning all interference levels; INT8, often assumed safe, also carries a smaller but real 
penalty in two of three models. The effect is specific to semantically similar (word-type) distractors and reverses sign 
under a numeric control condition, and is mechanistically linked to a rise in same-key intrusion errors under 
INT4 (21.5\% $\to$ 24.6\% of trials, $p=4.8\times10^{-7}$). A follow-up ablation shows the effect originates in the quantized 
transformer backbone rather than the output projection layer. These results suggest that bitsandbytes 4-bit quantization can impose an additional cost on applications relying on long, updatable, semantically dense contexts, even when aggregate benchmark accuracy appears largely unaffected.

We release our code and tokenizer-verified vocabulary construction method at \url{https://github.com/ShayanShahrabi/compress-and-forget}
\end{abstract}

\section{Introduction}

Post-training quantization (PTQ) is the default path by which most
practitioners deploy open-weight LLMs. 8-bit and 4-bit weight quantization
via libraries such as bitsandbytes \citep{dettmers2022llmint8,
dettmers2023qlora}, GPTQ \citep{frantar2023gptq}, and AWQ
\citep{lin2024awq} routinely report negligible loss on aggregate benchmarks
(perplexity, MMLU, standard QA). This has produced a general
assumption that 4-bit quantization is \emph{safe enough} for most purposes.

Separately, a growing line of work shows that LLMs are systematically
vulnerable to \emph{proactive interference} (PI); when a value is updated
multiple times within a context (for example, a meeting time that is first
set to 2pm, then moved to 3pm, then moved again to 4pm), retrieval of the
final value degrades log-linearly as the number of prior updates
grows, even though the correct value is the one nearest the query
\citep{wang2025unable}. This
failure mode is distinct from long-context retrieval failure in the
\emph{needle-in-a-haystack} sense \citep{liu2023lost}: the interfering values are
short, few in number relative to full context windows, and the task
provides no ambiguity about which value is being asked for. A recent
large-scale study across 39 models further shows that PI is a stronger and
more universal failure mode in transformers than the symmetric case
(retroactive interference), and that it reflects a distinct,
capacity-independent mechanism -- a bias toward protecting early
information at the expense of recent updates \citep{chattaraj2026transformers}.

These two lines of work have not been connected. It is not known whether
the aggregate-benchmark robustness of quantized models extends to
interference-heavy retrieval, or whether compressing model weights degrades
this specific, already-fragile capability disproportionately. This matters
practically, where applications with long-lived, frequently updated states
(multi-turn assistants, document revision tracking, dialogue systems that
track user preferences) are exactly the setting where PI is most likely to
occur, and are also the setting most likely to be served by quantized
models for cost reasons.

\begin{figure}
    \centering
    \includegraphics[width=1\linewidth]{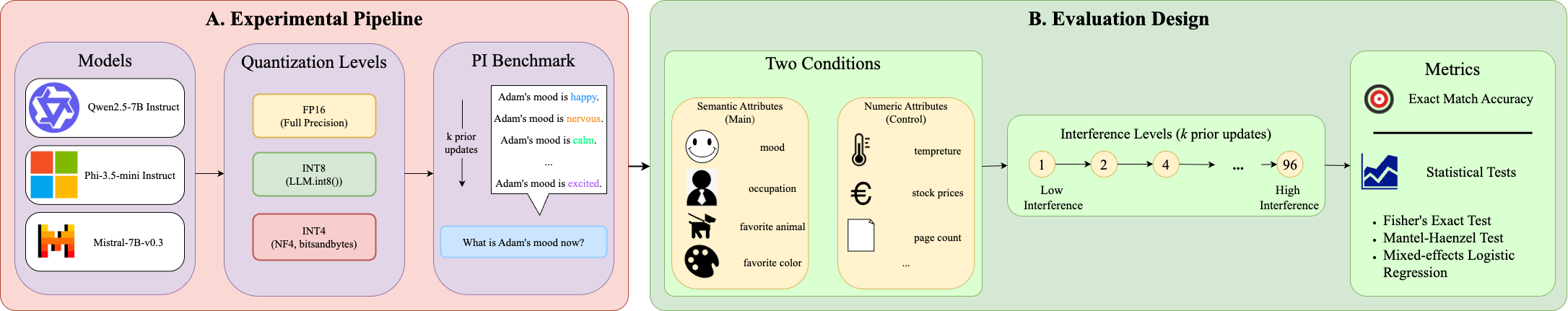}
    \caption{Overview of the experimental pipeline and evaluation design.
    (A) Each of three instruction-tuned models is loaded at three
    precisions (FP16, INT8, INT4/NF4 via bitsandbytes) and evaluated on
    the PI-LLM retrieval benchmark, in which a subject's attribute is
    overwritten $k$ times and the model is queried for only the most
    recent value. (B) Two attribute conditions are tested in parallel --
    semantic (word-type) attributes as the main condition and numeric
    attributes as a control -- across interference levels from $k=1$ to
    $k=96$, with exact-match accuracy and paired statistical tests
    (Fisher's exact, Mantel--Haenszel, mixed-effects logistic regression)
    computed at each level.}
    \label{fig:overview}
\end{figure}

We test this directly. Holding the PI-LLM retrieval task fixed, we vary
only quantization precision (FP16 / INT8 / INT4) across three open,
architecturally distinct 4--7B instruction-tuned models. We find a
consistent, statistically significant effect: 4-bit quantization
specifically degrades retrieval under semantic interference (e.g.
mood, occupation, favorite color/animal attributes), while an arbitrary
numeric-attribute control condition shows no such effect. This specificity
is the central finding of the research. It shows that quantization is not simply \textit{noisier} in a
way that hurts all retrieval equally -- it selectively erodes the
mechanism that resists semantically confusable distractors. Figure \ref{fig:overview} present an overview of what the procedure looks like.

Our main contributions are summarized as follows:

\begin{itemize}
  \item We provide an initial systematic evaluation of how post-training quantization affects proactive interference in LLMs, examining three precision levels across three architecturally distinct open-source models.
  \item We show the effect is specific to semantic (word-type)
  interference and does not generalize to arbitrary (numeric) interference,
  ruling out a generic noise-injection account.
  \item We identify a \textit{cliff pattern} rather than a graceful degradation:
  INT8 is statistically indistinguishable from FP16 in two of three models,
  while INT4 shows a significant, consistent drop -- suggesting the effect
  is not simply proportional to quantization aggressiveness.
  \item We release our task-generation code, tokenizer-verified vocabulary
  construction method, and full trial-level results.
\end{itemize}

\section{Related Work}

\paragraph{Proactive interference in LLMs.}
\citet{wang2025unable} introduce PI-LLM, adapting the classical proactive
interference paradigm from cognitive psychology: a key is
repeatedly rebound to new values within a context, and the model is queried
for only the final value. Despite the target being unambiguous and recency-
adjacent, retrieval accuracy declines log-linearly as the number of prior
rebindings grows; the authors show this holds across model families and is
not resolved by prompting the model to ignore earlier values, nor
reliably by chain-of-thought reasoning. \citet{chattaraj2026transformers}
extend this by directly contrasting proactive and retroactive interference
across 39 models, finding PI dominates RI universally (Cohen's $d=1.73$)
-- the opposite of the typical human pattern -- and that PI resistance is
uncorrelated with model scale, unlike RI resistance. This suggests PI
reflects a structural property of transformer attention (a primacy bias)
rather than a capacity limitation that scales away with larger models.
\citet{learningtoforget2026} propose a sleep-inspired memory consolidation
mechanism to mitigate PI, framing it as an learned, selective-forgetting
problem analogous to biological memory consolidation. None of this line of
work examines quantization as a variable.

\paragraph{Post-training quantization.}
LLM.int8() \citep{dettmers2022llmint8} and QLoRA \citep{dettmers2023qlora}
established that 8-bit and 4-bit (NF4) post-training quantization preserve
aggregate task performance with minimal loss. GPTQ \citep{frantar2023gptq}
and AWQ \citep{lin2024awq} improve on accuracy-per-bit through more
sophisticated calibration, but the underlying evaluation methodology in
this literature is near-universally aggregate benchmark accuracy (perplexity,
MMLU-style QA), not fine-grained behavioral failure modes. A recent
large-scale benchmarking effort, PTQ-Bench \citep{zhao2025ptqbench},
systematically taxonomizes weight-only PTQ strategies (compensation-based,
optimization-based, rotation-based, and salience-based) and evaluates their
cross-bitwidth, cross-architecture, and cross-modality robustness across
model sizes from 7B to 70B; notably, its recommendations are again derived
entirely from aggregate task metrics rather than from behavioral probes of
a specific cognitive-style failure mode such as PI, and it does not
evaluate bitsandbytes-style quantization specifically. This gap -- broad,
systematic PTQ benchmarking that nonetheless stays at the aggregate-metric
level -- is precisely what motivates our narrower but behaviorally-targeted
evaluation. Separately,
\citet{zhang2025whenless} study quantization's effect on \emph{catastrophic
forgetting during continual fine-tuning}, finding -- counterintuitively --
that 8-bit quantization noise acts as an implicit regularizer that
\emph{improves} retention relative to FP16, while 4-bit quantization is
more sensitive to replay buffer size. This is a training-time phenomenon
(weights are updated sequentially across tasks); our work is purely
inference-time (weights are frozen, only in-context retrieval is tested),
making the two effects mechanistically distinct despite superficial
similarity in framing. \citet{zhang2024catastrophic} show a
related but distinct interaction: quantization can causally resurface
knowledge that fine-tuning-based unlearning had suppressed, again pointing
to quantization interacting with a model's information-handling behavior
in ways aggregate benchmarks do not surface. Outside the language-model
setting, \citet{fu2026quantwm} present an empirical study of quantization
in visual world models (DINO-WM), showing that quantization-induced
failures in planning tasks are task-specific and asymmetric across model
components (encoder vs.\ predictor) rather than a simple function of
bitwidth. While this work evaluates a different model class and modality
entirely, it lends independent, cross-domain support to a theme our
results also suggest: aggregate bitwidth-vs-accuracy summaries can obscure
component- or task-specific quantization failure modes that only appear
under targeted behavioral evaluation.

\paragraph{Long-context retrieval.}
\citet{liu2023lost} show LLMs exhibit position-dependent retrieval accuracy
over long contexts (lost in the middle), a distinct phenomenon from PI:
their setting involves a single unambiguous target among irrelevant
distractors at varying positions, not a repeatedly overwritten value. Our
prompts are short relative to typical long-context evaluations (at most a
few hundred tokens even at our highest interference level), so our effect
is not attributable to position-in-context retrieval degradation of the
kind Liu et al.\ document. More directly relevant, \citet{mekala2025quantization}
systematically evaluate several post-training quantization methods on
long-context tasks and report that low-bit weight-only quantization,
including bitsandbytes' NF4 scheme, is disproportionately damaging in
retrieval-heavy long-context settings compared to other 4-bit quantizers
tested. Although their tasks involve much longer contexts and a different
retrieval paradigm (no repeated key-rebinding) than ours, their finding
that NF4 specifically -- rather than 4-bit quantization in general -- is
among the worst-performing methods for retrieval-heavy tasks converges
with our own restriction to bitsandbytes and strengthens the case that the
effect we report is not an idiosyncrasy of our particular task design.

\section{Method}

This section gives a brief overview of the task, models, and analysis
approach; full implementation detail (exact prompts, scoring rules,
vocabulary construction, reproducibility configuration, trial budgets, and
statistical formulas) is deferred to the Appendix section.

\subsection{Task design}
\label{sec:task-design}

We adapt the PI-LLM key-rebinding paradigm from \citep{wang2025unable} where a named
subject's attribute is updated $k$ times in sequence, defining the
interference level, and the model must retrieve only the final
value, disregarding the $k-1$ overwritten prior values. We test eight
attributes split into two conditions across interference levels
$k \in \{1,2,4,8,16,32,64,96\}$. Four \textbf{word-type (semantic)}
attributes (mood, favorite color, favorite animal, occupation), where distractors are
drawn from a shared semantic category, and three \textbf{numeric
(control)} attributes (temperature, stock price,
page count), where distractors are arbitrary integers with
minimal semantic confusability. Prompts are passed through each model's
native chat template as a single user message with no system prompt, and
models are instructed to answer with a single word or number. Exact
prompt templates, scoring criteria, and our tokenizer-verified vocabulary
construction procedure are given in Appendix~\ref{app:task-details}.

\subsection{Models and quantization}
\label{sec:models}

We evaluate three open-weight, instruction-tuned models spanning distinct
training organizations and architectures: Qwen2.5-7B-Instruct
\citep{qwen2024}, Mistral-7B-Instruct-v0.3 \citep{jiang2023mistral}, and
Phi-3.5-mini-instruct \citep{abdin2024phi3}. Each is loaded at three
precisions via the bitsandbytes library \citep{dettmers2022llmint8,
dettmers2023qlora}: FP16 (native), INT8 (default LLM.int8() decomposition),
and INT4 (NF4 with double quantization). No layers were explicitly excluded from quantization for any arm,
including the language modeling head (\texttt{lm\_head}). A follow-up
ablation (Section~\ref{sec:lmhead-ablation}) found, however, that under
this library version, the default INT4 loading path leaves lm\_head at
full precision regardless -- confirmed by inspecting the loaded module
directly. All INT4 results reported in this paper are
therefore \emph{backbone-only} quantization; Section~\ref{sec:lmhead-ablation}
shows this does not affect our conclusions. Generation is greedy throughout
(deterministic given a prompt); the only source of trial-to-trial
randomness is the sampled subject name and distractor values, controlled by
an explicit random seed. Exact loading configuration and package versions,
provided for reproducibility, are given in Appendix~\ref{app:repro}.

\subsection{Trial design and statistical analysis}
\label{sec:stats}

For a given random seed, the exact sequence of trials -- subject,
distractors, order -- is identical across all three quantization levels,
making every FP16-vs-INT4 (or FP16-vs-INT8) comparison a paired comparison
at the trial level (verified with zero mismatches across 23{,}075 unique
trials). The exact per-level trial budget and further detail on the paired
design are given in Appendix~\ref{app:trials}.

We report McNemar's exact test as our primary significance test,
exploiting this trial-level pairing, at each model's most-powered
interference level (selected for being farthest from ceiling/floor
effects); we retain the unpaired Fisher's exact test and a pooled
Mantel--Haenszel test as more conservative cross-checks. To confirm the
effect is not an artifact of level selection, we additionally fit a
mixed-effects logistic regression using all eight interference levels
simultaneously, with random intercepts for model, seed, and attribute, and
a random-slopes robustness variant. We separately classify every incorrect
word-type response as a same-key intrusion or non-matching error, and
stratify the numeric-attribute comparison by token length to rule out a
tokenization confound. Full formulas, model specifications, and robustness
details are given in Appendix~\ref{app:stats-detail}.

\section{Results}

\subsection{Overall accuracy}
\label{sec:overall-accuracy}

Table \ref{tab:overall-accuracy} shows overall accuracy (all attributes,
all interference levels pooled) by model and quantization. Aggregate
accuracy differences between FP16 and INT4 are small in absolute terms
(1.1--2.4 percentage points) -- consistent with the standard finding in the
quantization literature that aggregate benchmarks are largely insensitive
to 4-bit quantization. This is precisely why the effect we report below
would not be visible from aggregate accuracy alone: it is concentrated in a
specific, high-interference regime that aggregate metrics average away.

\begin{table}[t]
\centering
\caption{Overall accuracy (all attributes and interference levels pooled) by model and quantization.}
\label{tab:overall-accuracy}
\begin{tabular}{lccc}
\toprule
Model & FP16 & INT8 & INT4 \\
\midrule
Qwen2.5-7B-Instruct        & 97.6\% & 97.6\% & 96.5\% \\
Mistral-7B-Instruct-v0.3   & 75.6\% & 74.4\% & 72.4\% \\
Phi-3.5-mini-instruct      & 57.2\% & 53.0\% & 59.5\% \\
\bottomrule
\end{tabular}
\end{table}

\subsection{Word-type accuracy declines sharply under INT4 at high interference}
\label{sec:word-decline}

\begin{figure}[t]
  \centering
  \includegraphics[width=\textwidth]{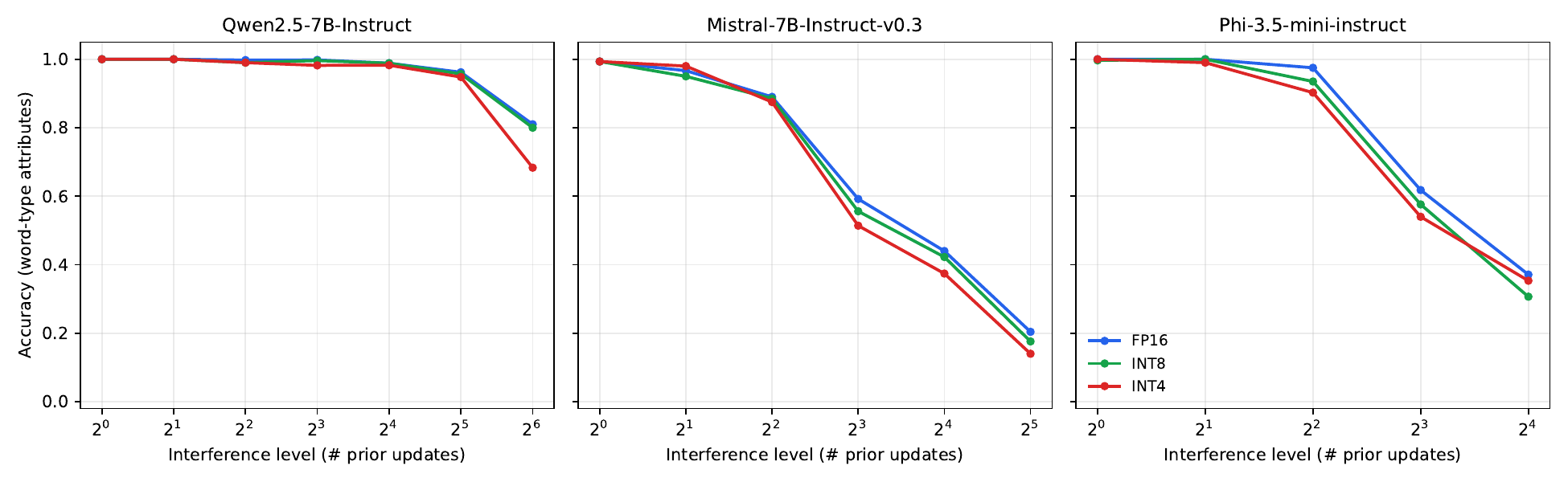}
  \caption{Word-type (semantic interference) accuracy vs.\ interference
  level, by model and quantization precision. Shaded region / error bars:
  omitted for clarity in the line plot; see Table \ref{tab:key-results} for
  exact accuracy and significance at each model's key level.}
  \label{fig:accuracy-by-level}
\end{figure}

Figure \ref{fig:accuracy-by-level} shows word-type accuracy as a function
of interference level. All three models show the expected log-linear-style decline in
accuracy as interference increases \citep{wang2025unable}; critically, the
INT4 curve separates visibly from FP16/INT8 at high interference in every
model.

Table \ref{tab:key-results} reports the quantitative comparison at each
model's most-powered interference level (Section~\ref{sec:stats}), using
McNemar's paired test (our primary test; Section~\ref{sec:stats}) on the
discordant trial pairs.

\begin{table}[t]
\centering
\caption{FP16 vs.\ INT4 accuracy at each model's most-powered
high-interference level (word-type attributes), with McNemar's paired
test. $n_{10}$ = trials correct under FP16 but not INT4; $n_{01}$ = the
reverse; OR $=n_{10}/n_{01}$. $n$ = paired trials (pooled across 5 seeds).
All three models show the same direction of effect in every individual
seed (15/15 seed-comparisons). The unpaired Fisher's exact test on the
same cells agrees in direction and significance but is markedly less
powered ($p=0.0005$, $0.0012$, $0.015$ respectively), as expected once the
trial-level pairing is discarded.}
\label{tab:key-results}
\begin{tabular}{lccccccc}
\toprule
Model & Level & FP16 & INT4 & $n_{10}$ / $n_{01}$ & OR & $p$ (McNemar) \\
\midrule
Qwen2.5-7B-Instruct      & 64 & 81.0\% & 68.3\% & 40 / 2   & 20.00 & $4.1\times10^{-10}$ \\
Mistral-7B-Instruct-v0.3 & 16 & 44.0\% & 37.4\% & 129 / 50 & 2.58  & $5.5\times10^{-9}$  \\
Phi-3.5-mini-instruct    & 8  & 61.8\% & 54.0\% & 54 / 15  & 3.60  & $2.6\times10^{-6}$  \\
\bottomrule
\end{tabular}
\end{table}

\textbf{Pooled unpaired cross-check.} A Mantel--Haenszel test combining
all three models' unpaired $2\times2$ tables yields a common odds ratio of
$1.40$ (95\% CI $[1.23, 1.59]$) and $p = 3.5\times10^{-7}$. This is
consistent with, but considerably more conservative than, the paired
McNemar result above, exactly as expected: pooling ignores the trial-level
pairing entirely, so it represents a lower bound on how confident the
paired design actually lets us be.

\textbf{Revised INT8 picture: paired testing reveals a real, smaller
effect in two of three models.} Table \ref{tab:int8-comparison} repeats
the same paired analysis for FP16 vs.\ INT8. Under the unpaired Fisher's
test (as originally reported), all three models showed confidence
intervals crossing 1 ($p=0.837$, $0.410$, $0.197$), suggesting INT8 was
statistically indistinguishable from FP16 -- a clean "cliff" pattern
(FP16 $\approx$ INT8 $\gg$ INT4). McNemar's paired test tells a different
story: Qwen remains a clean FP16 $\approx$ INT8 case ($p=0.629$), but
Mistral and Phi both show a real, if smaller-magnitude, INT8 accuracy
penalty ($p=0.018$ and $p=0.004$ respectively) that the unpaired test
lacked the power to detect. We take this as the more trustworthy result,
since it is the appropriately-matched test for our paired design. The
practical picture is therefore not a clean cliff but a graded one: INT4
carries a large, robust penalty in every model (OR $2.6$--$20.0$); INT8
carries a smaller but still real penalty in two of three models
(OR $1.8$--$2.5$), with Qwen as the one model where INT8 appears safe.
Phi additionally shows one non-monotonic condition at $k=16$, reported
rather than omitted in Appendix~\ref{app:additional-results}. INT8 should
therefore not be assumed safe by default for every model.

\begin{table}[t]
\centering
\caption{FP16 vs.\ INT8 accuracy at the same levels as Table
\ref{tab:key-results} (word-type attributes), McNemar's paired test. The
unpaired Fisher's exact test at these same cells gives $p=0.837$, $0.410$,
$0.197$ respectively (all n.s.) -- the paired test recovers a real effect
in two of the three models that the unpaired test misses.}
\label{tab:int8-comparison}
\begin{tabular}{lccccccc}
\toprule
Model & Level & FP16 & INT8 & $n_{10}$ / $n_{01}$ & OR & $p$ (McNemar) \\
\midrule
Qwen2.5-7B-Instruct      & 64 & 81.0\% & 80.0\% & 10 / 7  & 1.43 & $0.629$ \\
Mistral-7B-Instruct-v0.3 & 16 & 44.0\% & 42.2\% & 46 / 25 & 1.84 & $0.018$ \\
Phi-3.5-mini-instruct    & 8  & 61.8\% & 57.6\% & 35 / 14 & 2.50 & $0.004$ \\
\bottomrule
\end{tabular}
\end{table}

\subsection{A standardized summary metric: Interference Endurance Score (IES)}
\label{sec:ies}

To enable standardized comparison with the PI-LLM literature and to
summarize the quantization penalty in a single number, we compute the
Interference Endurance Score (IES; \citealp{wang2025unable}): the area
under the accuracy-vs-interference-level curve, integrated over a
log$_2$-scaled interference axis. We report a normalized variant, dividing
by the tested log$_2$ range, so IES falls on the same $[0,1]$ scale as
accuracy and remains comparable across models even though Mistral and
Phi's smaller tokenizer-filtered word-type vocabularies cap their maximum
tested interference level below Qwen's (Table~\ref{tab:vocab-sizes}). IES
is computed directly from the existing trial-level dataset; no new
experiments were required.

Table~\ref{tab:ies-results} reports IES and the resulting quantization
penalty for each model. INT4 quantization reduces IES relative to FP16 by
1.8\% (Qwen2.5-7B), 5.1\% (Mistral-7B), and 5.2\% (Phi-3.5-mini) on
word-type attributes, with INT8 falling consistently between FP16 and
INT4 in every model -- monotonic in the expected direction
(FP16 $\ge$ INT8 $\ge$ INT4).

\begin{table}[t]
\centering
\caption{Interference Endurance Score (IES; area under the
accuracy-vs-log$_2$-interference-level curve, normalized to $[0,1]$) by
model and quantization, word-type attributes. \% lost is relative to
FP16. Tested interference-level ranges differ by model due to the
vocabulary-size cap in Table~\ref{tab:vocab-sizes}.}
\label{tab:ies-results}
\begin{tabular}{lccccc}
\toprule
Model & Levels tested & FP16 & INT8 & INT4 & INT4 \% IES lost \\
\midrule
Qwen2.5-7B-Instruct      & 1--64 (7 levels) & 0.975 & 0.972 & 0.957 & 1.82\% \\
Mistral-7B-Instruct-v0.3 & 1--32 (6 levels) & 0.698 & 0.680 & 0.662 & 5.09\% \\
Phi-3.5-mini-instruct    & 1--16 (5 levels) & 0.820 & 0.791 & 0.777 & 5.16\% \\
\bottomrule
\end{tabular}
\end{table}

\paragraph{IES understates, rather than contradicts, the paper's cliff
finding for Qwen.} Qwen's accuracy remains near ceiling ($\ge$96\%)
through interference level 32 under every precision and only collapses
sharply at level 64 (Table~\ref{tab:key-results}: 81.0\% $\to$ 68.3\%, a
13-point drop). Because IES averages over the full tested range and most
of that range is saturated and identical across quantization levels for
Qwen, this large, practically important local effect is diluted to a
1.8\% aggregate IES loss. We therefore report IES alongside, not in place
of, the per-level cliff analysis (Table~\ref{tab:key-results}): Mistral
and Phi show more gradual decline starting at lower interference levels,
so their IES penalties more directly reflect the accuracy curves in
Figure~\ref{fig:accuracy-by-level}, while Qwen's small IES penalty
specifically should not be read as evidence against its cliff.

\begin{figure}[t]
  \centering
  \includegraphics[width=\textwidth]{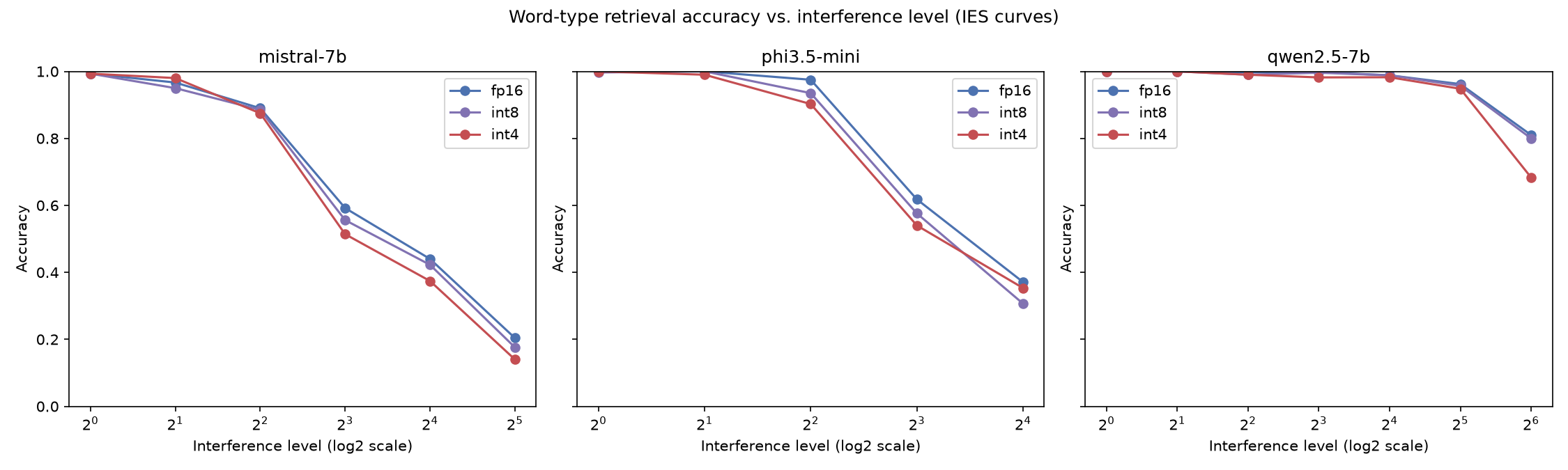}
  \caption{Accuracy vs.\ log$_2$-scaled interference level, FP16/INT8/INT4
  overlaid per model -- the curve underlying the IES computation in
  Table~\ref{tab:ies-results}. Illustrates the cliff-dilution effect
  discussed in the text: Qwen's near-flat curve through level 32 versus
  Mistral's and Phi's more gradual declines.}
  \label{fig:ies-curves}
\end{figure}

Word-vs-numeric IES comparisons (Appendix~\ref{app:additional-results})
partially corroborate the specificity result of
Section~\ref{sec:specificity}: Mistral's INT4 penalty is larger under
word-type interference (5.1\%) than numeric (2.6\%), consistent with the
paper's central claim, though the pattern is not uniform across all three
models.

\subsection{Mixed-effects regression confirms the effect using all interference levels}
\label{sec:mixed-model}

The per-model analysis above necessarily selects one level per model for
statistical power. To confirm this is not an artifact of that selection,
Table \ref{tab:mixed-model} reports the mixed-effects logistic regression
described in Section~\ref{sec:stats}, fit separately on word-type and
numeric trials, using \emph{all} eight interference levels and all three
models simultaneously.

\begin{table}[t]
\centering
\caption{Mixed-effects logistic regression: $\text{correct} \sim
\text{quant\_mode} * \log_2(\text{level}) + (1\mid\text{model}) +
(1\mid\text{seed}) + (1\mid\text{attribute})$. FP16 is the reference level.
The key row in each panel is the INT4 $\times$ log-interference-level
interaction: negative for word-type, positive for numeric -- opposite
signs, both highly significant, even after accounting for attribute-level
difficulty heterogeneity via the added random intercept.}
\label{tab:mixed-model}
\begin{tabular}{lcccc}
\toprule
Term & Coef.\ ($\beta$) & OR & 95\% CI & $z$ ($p$) \\
\midrule
\multicolumn{5}{l}{\textit{Word-type attributes (main hypothesis)}} \\
\midrule
int8 (main effect)             & $-0.256$ & 0.77 & [0.73, 0.82] & $-7.93$ ($2.2\times10^{-15}$) \\
int4 (main effect)             & $-0.143$ & 0.87 & [0.81, 0.92] & $-4.44$ ($9.1\times10^{-6}$) \\
int8 $\times \log_2(\text{level})$ & $+0.023$ & 1.02 & [1.01, 1.04] & $2.73$ ($0.0063$) \\
\textbf{int4 $\times \log_2(\text{level})$} & $\mathbf{-0.046}$ & \textbf{0.96} & \textbf{[0.94, 0.97]} & $\mathbf{-5.43}$ ($\mathbf{5.7\times10^{-8}}$) \\
\midrule
\multicolumn{5}{l}{\textit{Numeric attributes (specificity control)}} \\
\midrule
int8 (main effect)             & $-0.213$ & 0.81 & [0.77, 0.85] & $-8.52$ ($<10^{-16}$) \\
int4 (main effect)             & $-0.190$ & 0.83 & [0.79, 0.87] & $-7.45$ ($9.2\times10^{-14}$) \\
int8 $\times \log_2(\text{level})$ & $+0.013$ & 1.01 & [1.00, 1.02] & $2.81$ ($0.0049$) \\
\textbf{int4 $\times \log_2(\text{level})$} & $\mathbf{+0.053}$ & \textbf{1.05} & \textbf{[1.04, 1.06]} & $\mathbf{11.05}$ ($\mathbf{<10^{-16}}$) \\
\bottomrule
\end{tabular}
\end{table}

For word-type attributes, the INT4 $\times$ log-interference-level
interaction is negative and highly significant: the INT4 accuracy penalty
relative to FP16 grows as interference increases. For numeric attributes,
the same interaction term is \emph{positive} and equally significant --
INT4's relative accuracy, if anything, improves slightly as numeric
interference increases. This is a stronger specificity result than "no
effect on the control condition": the effect runs in the opposite
direction, which is difficult to explain under any account where
quantization simply injects generic noise into retrieval. This pattern is
robust to adding random slopes for interference by model and attribute,
and the main-effect coefficients (which describe the quantization gap only
at the lowest interference level, $k=1$, near ceiling for all conditions)
should not be read as practically meaningful on their own; both points are
detailed further in Appendix~\ref{app:additional-results}.

\subsection{Specificity: opposite-signed effect on numeric (non-semantic) interference}
\label{sec:specificity}

\begin{figure}[t]
  \centering
  \includegraphics[width=\textwidth]{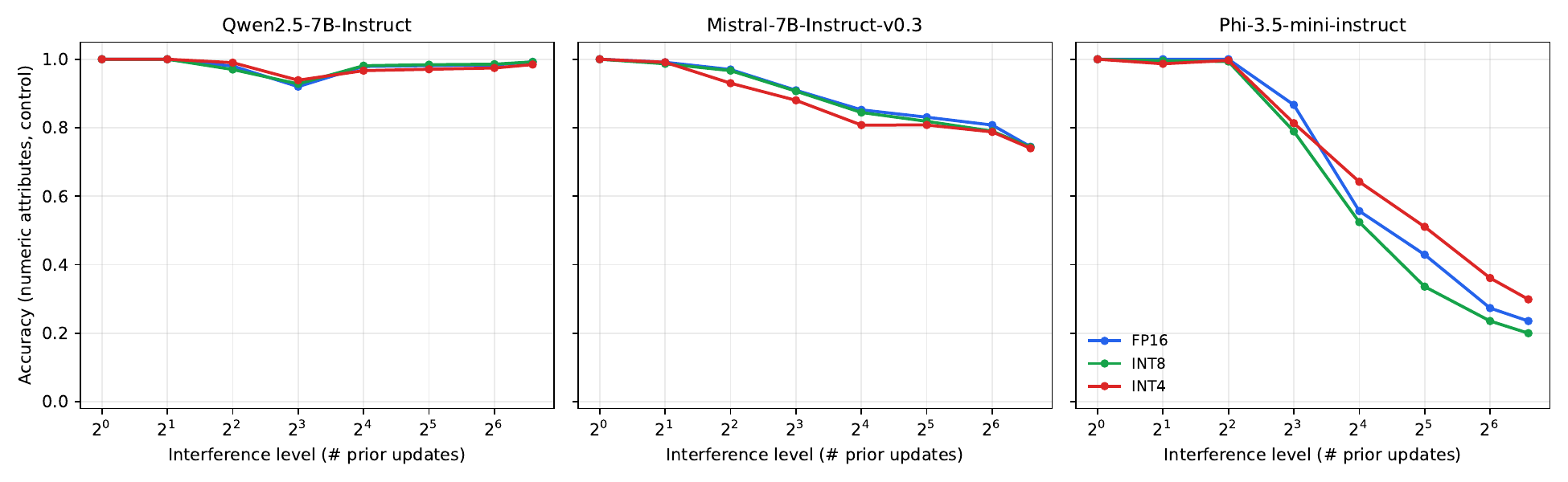}
  \caption{Numeric-attribute (control) accuracy vs.\ interference level, by
  model and quantization precision. Baseline interference is present, but
  no consistent quantization effect emerges at any level, in contrast to
  Figure \ref{fig:accuracy-by-level}.}
  \label{fig:numeric-control}
\end{figure}

Figure \ref{fig:numeric-control} shows the numeric-attribute (control)
accuracy-by-level curves. Baseline interference is present here too: pooled
across all three models, accuracy declines from $\sim$100\% to
$\sim$65--70\% as $k$ grows to its highest tested values, confirming the
task design elicits genuine interference-driven decline even for arbitrary
values. This pooled figure, however, masks substantial per-model
heterogeneity: Qwen remains comparatively robust even at high $k$ (numeric
accuracy stays above 90\% through most of the tested range), while Mistral
and Phi decline much more steeply. This is exactly why the mixed-effects
model (Section~\ref{sec:mixed-model}) includes a random intercept for
model: it separates each model's overall difficulty level from the
quantization-by-interference interaction we are actually testing.

At the three highest numeric interference levels (32, 64, 96, pooled
across all three models), FP16 vs.\ INT4 differences are individually not
significant ($p=0.225$, $p=0.123$, $p=0.194$ respectively) and not
consistently signed in FP16's favor -- consistent with, though less
precise than, the mixed model's positive interaction coefficient. Together,
these results support the paper's central claim: 4-bit quantization does
not generically add retrieval noise that would degrade any
interference-heavy task equally. It selectively erodes whatever mechanism
resists \emph{semantically confusable} distractors, while leaving
resistance to arbitrary, non-confusable distractors intact -- and if
anything behaves oppositely there. Because the numeric vocabulary is not
tokenizer-filtered to be single-token, unlike the word-type pools, we
separately verified that token-length composition is flat across
interference levels and that the specificity result survives stratifying
by token length (Appendix~\ref{app:additional-results}), ruling out a
tokenization confound as the explanation.

\begin{figure}[t]
  \centering
  \includegraphics[width=0.8\textwidth]{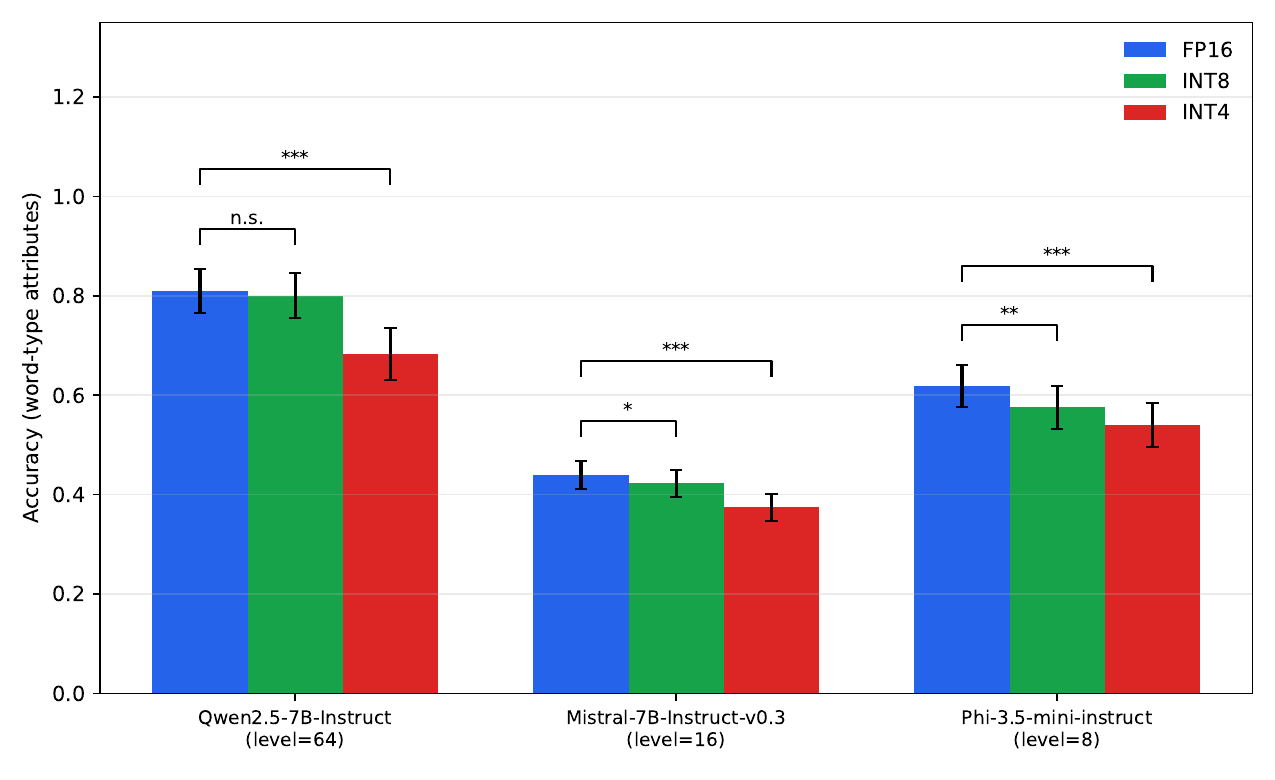}
  \caption{Accuracy at each model's key high-interference level, word-type
  attributes only, with 95\% confidence intervals. Brackets denote
  significance via McNemar's paired test (our primary test,
  Section~\ref{sec:stats}): the lower bracket compares FP16 vs.\ INT8, the
  upper bracket FP16 vs.\ INT4. *** $p<0.001$, ** $p<0.01$, * $p<0.05$,
  n.s.\ not significant. Note the INT8 bracket is now significant for
  Mistral and Phi -- the paired test recovers an effect the unpaired
  Fisher's test (used in an earlier version of this analysis) missed.}
  \label{fig:key-level-bars}
\end{figure}

\subsection{Mechanism: quantization increases the rate of same-key intrusions}
\label{sec:intrusion}

The results above establish that INT4 quantization lowers retrieval
accuracy under semantic interference, but accuracy alone does not show
\emph{how} the model fails. We reconstruct the exact ordered sequence of
overwritten distractor values for every trial (deterministic given the
seed; see Appendix~\ref{app:stats-detail}) and classify every incorrect
word-type response as either a \emph{same-key intrusion} -- the extracted
answer exactly matches one of the $k-1$ earlier, overwritten values for
that trial -- or a non-matching error.

The overwhelming majority of errors are same-key intrusions at every
precision level: 99.9\% at FP16 (2{,}007/2{,}009 errors), 99.9\% at INT8
(2{,}167/2{,}170), and 99.8\% at INT4 (2{,}297/2{,}302). This confirms
that when these models fail on this task, they are not answering at
random -- they are retrieving a real but stale value, consistent with the
PI-LLM literature's framing of this failure mode \citep{wang2025unable}.

\begin{figure}[t]
  \centering
  \includegraphics[width=0.55\textwidth]{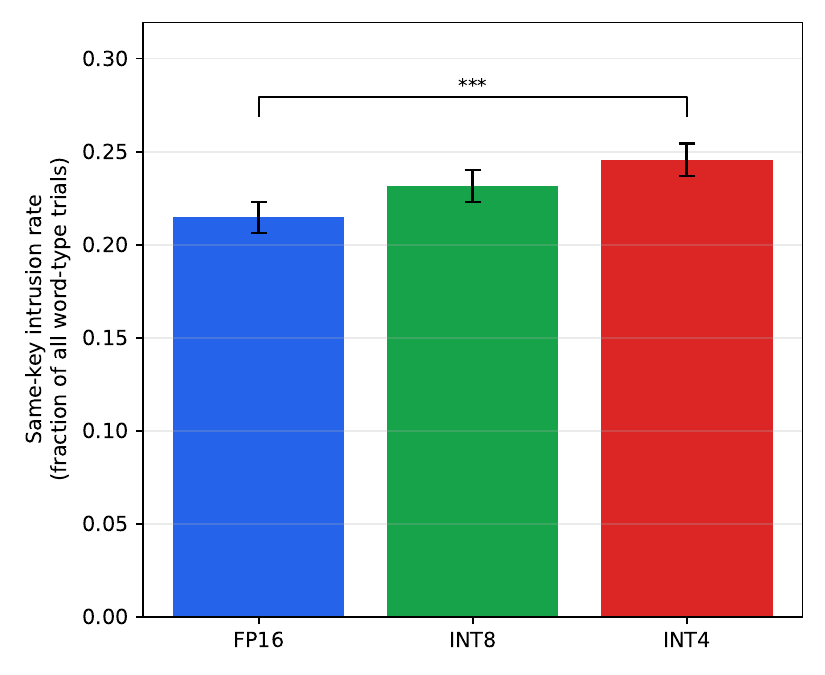}
  \caption{Same-key intrusion rate (fraction of all word-type trials whose
  wrong answer exactly matches an earlier overwritten value), by
  quantization precision, pooled across all three models, seeds,
  attributes, and interference levels. Error bars: 95\% CI. Bracket:
  McNemar's paired test, FP16 vs.\ INT4 ($p=4.8\times10^{-7}$).}
  \label{fig:intrusion-rate}
\end{figure}

Critically, the \emph{rate} of same-key intrusions, expressed as a
fraction of \emph{all} trials (not just errors), rises monotonically and
significantly with more aggressive quantization: 21.5\% of all word-type
trials at FP16, 23.2\% at INT8, and 24.6\% at INT4 (Figure
\ref{fig:intrusion-rate}). A paired McNemar's test on the FP16-vs-INT4
intrusion indicator, pooled across all models, seeds, attributes, and
interference levels, confirms this is significant (OR $=0.84$, 95\% CI
$[0.78, 0.90]$, $p=4.8\times10^{-7}$). Intrusion \emph{recency} -- how many
positions back the intruding value came from -- is comparable across
precision levels (mean position 13.8 at FP16, 13.2 at INT8, 13.2 at INT4),
suggesting quantization primarily changes \emph{how often} the model
confuses an earlier value for the current one, not \emph{which} earlier
value it confuses.

This is direct behavioral evidence for the mechanism this paper proposes
in Section~\ref{sec:discussion}: quantization does not add generic answer
noise, it specifically increases the rate at which semantically similar
candidate values are confused with one another, which is exactly the
signature that a coarsening of representational precision among
semantically similar candidates -- rather than a general accuracy loss --
would predict.

\subsection{Isolating the contribution of the output projection (lm\_head)}
\label{sec:lmhead-ablation}

A reviewer asked how much of the INT4 effect reported above is
attributable to quantizing the output/vocabulary projection (lm\_head)
specifically, versus the transformer backbone. We
address this with a follow-up ablation, restricted to word-type
attributes at each model's key high-interference level, using the same 5
seeds and identical environment as the main experiments. New trials for
this ablation were generated to be paired by construction with the
existing dataset, though this pairing was not re-verified with the same
exhaustive, zero-mismatch check used for the main dataset; we treat this
ablation's paired comparisons accordingly as a well-motivated but slightly
less rigorously verified companion to the paper's primary paired results.

As noted in Section~\ref{sec:models}, our original INT4 condition did
not, in practice, quantize lm\_head at all, which required a
methodological correction before the intended comparison could be run; the
full story, including how this was diagnosed and corrected, is given in
Appendix~\ref{app:lmhead-narrative}. The corrected ablation condition
verifiably quantizes lm\_head on top of the existing backbone-only INT4
model and differs from it on 4.0\% of paired trials (203/5{,}050),
confirming the two are genuinely different models.

\paragraph{Results.} Table \ref{tab:lmhead-ablation} compares accuracy
under backbone-only INT4 (the condition used throughout the paper)
against INT4 with lm\_head also quantized, at each model's key
interference level. Additionally quantizing lm\_head produces no
statistically detectable change in accuracy in any of the three models
(paired McNemar's test, all three comparisons $p=1.0$, with only 3--4
discordant trials out of 300--1{,}200 paired trials in either direction).
Decomposing each model's total FP16$\to$INT4(+lm\_head)
accuracy gap into a backbone-attributable share and an
lm\_head-attributable share shows that quantizing lm\_head
accounts for at most a few percentage points of the total effect: 2.5\%
for Qwen2.5-7B, 1.3\% for Mistral-7B, and 0.0\% for Phi-3.5-mini, none of
which are statistically distinguishable from zero given the above. The
underlying FP16$\to$INT4 effect itself remains highly significant with
lm\_head quantized in every model (the lm\_head-quantized condition
vs.\ FP16: $p=8.7\times10^{-10}$, $3.2\times10^{-9}$, and
$3.8\times10^{-6}$ for Qwen, Mistral, and Phi respectively).

\begin{table}[t]
\centering
\caption{Effect of additionally quantizing lm\_head on top of the
backbone-only INT4 condition, word-type attributes, at each model's key
interference level (same levels as Table \ref{tab:key-results}). ``INT4
(backbone)'' is the condition used throughout the paper; ``INT4
(+lm\_head)'' is the corrected ablation condition with lm\_head also
quantized. $p$-values are from McNemar's paired test comparing the two
conditions directly; none reach significance.}
\label{tab:lmhead-ablation}
\small
\begin{tabular}{lcccccc}
\toprule
Model & Level & $n$ & \shortstack{INT4\\(backbone)} & \shortstack{INT4\\(+lm\_head)} & \shortstack{lm\_head\\share of gap} & $p$ \\
\midrule
Qwen2.5-7B-Instruct      & 64 & 300   & 68.3\% & 68.0\% & 2.5\% & $1.0$ \\
Mistral-7B-Instruct-v0.3 & 16 & 1{,}200 & 37.4\% & 37.3\% & 1.3\% & $1.0$ \\
Phi-3.5-mini-instruct    & 8  & 500   & 54.0\% & 54.0\% & 0.0\% & $1.0$ \\
\bottomrule
\end{tabular}
\end{table}

\paragraph{Interpretation.} Across all three models, this ablation gives a
consistent answer: the proactive-interference penalty under INT4
quantization is backbone-driven, not an artifact of quantizing the output
projection. This is consistent with, and lends additional support to, the
representational-coarsening mechanism proposed in
Section~\ref{sec:discussion} and directly evidenced by the same-key
intrusion analysis (Section~\ref{sec:intrusion}): both point toward the
effect originating in the residual stream / backbone representations
rather than in the final projection to vocabulary logits.

\section{Discussion}
\label{sec:discussion}

Five findings together paint a consistent picture. First, the effect is
real and generalizes: all three architecturally distinct models show the
same direction of effect, every individual seed agrees with the aggregate
direction (15/15), and our primary paired test (McNemar's) is significant
in every model ($p \le 2.6\times10^{-6}$), with a more conservative
unpaired pooled test agreeing ($p=3.5\times10^{-7}$, common OR $1.40$,
95\% CI $[1.23, 1.59]$). Second, the effect is not an artifact of
selecting a favorable interference level per model, nor of
attribute-level difficulty differences, nor of unmodeled model- or
attribute-specific interference trends: a mixed-effects regression using
all eight levels, all three models, and a random intercept per attribute
confirms the INT4 penalty grows with interference
($\beta=-0.046$, $p=5.7\times10^{-8}$), and this survives essentially
unchanged when random slopes for interference by model and attribute are
added ($\beta=-0.048$, $p=1.9\times10^{-8}$). Third, the effect is
specific: it appears under semantic (word-type) interference and runs in
the \emph{opposite} direction under numeric interference
($\beta=+0.053$, $p<10^{-16}$), a result that survives stratifying by the
numeric values' token length (ruling out a tokenization confound as the
explanation) -- which together argue against a generic "quantization adds
noise" explanation and toward a more specific mechanism -- e.g.\ that
4-bit weight quantization coarsens the representational precision needed
to keep semantically similar candidate values distinct in the residual
stream, without similarly affecting representations of arbitrary,
mutually dissimilar numeric tokens. Fourth, this proposed mechanism is not
merely inferred from the accuracy pattern but directly observed in the
errors themselves: virtually all incorrect responses (99.8--99.9\%, at
every precision level) are same-key intrusions rather than unrelated
answers, and the \emph{rate} of these intrusions rises significantly under
INT4 relative to FP16 (24.6\% vs.\ 21.5\% of all trials,
$p=4.8\times10^{-7}$) -- direct behavioral evidence that quantization
specifically increases confusability among semantically similar
candidates, rather than merely adding generic answer noise. A follow-up
ablation isolating lm\_head (Section~\ref{sec:lmhead-ablation})
reinforces this same residual-stream account from a different angle:
additionally quantizing the output projection on top of the backbone adds
no statistically detectable effect in any of the three models (all
$p=1.0$), and accounts for at most 2.5\% of the total FP16$\to$INT4 gap
where lm\_head is quantized, so the effect is attributable to the
quantized backbone rather than the final vocabulary projection. Fifth, the
effect is graded rather than a clean cliff: our primary paired test shows
INT4 carries a large, robust penalty in every model (OR $2.6$--$20.0$),
while INT8 carries a smaller but still real penalty in two of three models
(Mistral, Phi; OR $1.8$--$2.5$) that an unpaired test lacks the power to
detect, with Qwen as the one model where INT8 remains statistically
indistinguishable from FP16. INT8 should therefore not be assumed uniformly
safe, though it remains meaningfully less risky than INT4 across the
board. This narrower, task- and precision-specific pattern is also
consistent with the broader theme identified by PTQ-Bench-style systematic
benchmarking \citep{zhao2025ptqbench} and by cross-domain evidence from
quantized world models \citep{fu2026quantwm}: quantization failure modes
are frequently task-specific and component-specific rather than a simple
monotonic function of bitwidth, so aggregate accuracy-vs-bitwidth
summaries -- even comprehensive ones -- can miss behaviorally important
failures that only targeted evaluation surfaces. A standardized summary
statistic (IES; Section~\ref{sec:ies}) corroborates this graded picture
while also illustrating its own limits: it confirms a real, monotonic
FP16 $\ge$ INT8 $\ge$ INT4 quantization penalty in every model, but
substantially understates Qwen's cliff specifically, since averaging over
the full interference range dilutes a large effect concentrated at a
single high interference level -- underscoring why we report both the
standardized metric and the per-level analysis rather than either alone.

\paragraph{Practical implications.} Applications that rely on 4-bit
quantized models to track frequently-updated, semantically dense state
(e.g.\ an assistant tracking a user's evolving preferences across a long
conversation, or a system tracking the current status of many similar
entities) may be more exposed to retrieval failures than aggregate
benchmark scores would suggest. INT8 quantization remains a meaningfully
safer choice than INT4 for such applications -- its accuracy penalty is
smaller in every model we tested, and undetectable in one of three -- but
our paired re-analysis indicates it is not risk-free either. Practitioners
should not treat INT8 as categorically safe by default for
interference-heavy retrieval tasks, particularly for models resembling
Mistral or Phi in this study, even though both quantization levels are
typically treated as "safe" compression levels in general-purpose
benchmarking.

\section{Limitations and Future Work}
\label{sec:limitations}

This work establishes a robust, multiply-verified effect within a specific
and well-controlled scope, which also defines the natural directions for
extending it. Our quantization coverage is bitsandbytes-specific
(LLM.int8/NF4); as Section~\ref{sec:lmhead-ablation} shows, this
concretely means the INT4 effect we report is a backbone effect, since
the library leaves lm\_head at full precision by default and a
forced-quantization ablation found no additional penalty from including
it. Extending this evaluation to calibration-based quantizers (GPTQ, AWQ)
and newer formats (EXL2, HQQ) -- several of which explicitly minimize
output divergence from the unquantized model and may handle lm\_head
differently by default -- would clarify whether the
effect is specific to bitsandbytes' NF4 scheme or reflects a more general
cost of aggressive weight compression; a PTQ-Bench-style
\citep{zhao2025ptqbench} sweep across quantizer families would be a
natural vehicle for this and was scoped for the present revision but not
yet completed. Activation and KV-cache quantization, increasingly common
in production serving stacks, are likewise promising directions we have
not yet tested.

Our numeric control condition uses a vocabulary that is not
tokenizer-filtered to be single-token, unlike the word-type pools; we
tested this directly (Section~\ref{sec:specificity}) and found token-length
composition flat across interference levels with the specificity result
holding in the dominant token-length stratum, which substantially
de-risks this design choice without fully replacing a prospectively
filtered design. Our prompts are also short-to-medium (at most a few
hundred tokens), so whether the effect strengthens or attenuates in the
8k+-token long-context regime is an open, practically important question
-- one that prior work on quantized long-context retrieval
\citep{mekala2025quantization} suggests could go either way and would be
worth testing directly with our paradigm.

The mechanism we propose -- that quantization coarsens representational
precision among semantically similar candidates -- is now supported by
two independent lines of behavioral evidence (the same-key intrusion
analysis of Section~\ref{sec:intrusion} and the backbone-localization
ablation of Section~\ref{sec:lmhead-ablation}), but neither constitutes a
representational-level test. Probing model internals directly (attention
patterns, residual-stream precision, activation distributions at the
layers implicated in candidate-value tracking) is a natural next step to
move this from a well-evidenced interpretation to a directly tested
claim. Finally, extending the model pool beyond the 3.8--7B range tested
here to larger and calibration-quantized-by-default models, and testing
whether chain-of-thought prompting or alternative decoding strategies
(only greedy decoding is tested here) interact with quantization, would
clarify how broadly this effect generalizes.

\section{Conclusion}

We show that bitsandbytes 4-bit post-training quantization significantly
degrades LLM retrieval accuracy under semantic proactive interference,
across three architecturally distinct open models, confirmed by
per-model McNemar's paired tests at high-power interference levels
(our primary test, exploiting the trial-level pairing built into the
experimental design), by a more conservative unpaired pooled test, and by
a mixed-effects regression using all interference levels simultaneously
-- with random intercepts for model, seed, and attribute, and, in a
robustness check, random slopes for interference by model and attribute
as well. This effect reverses sign under arbitrary (non-semantic)
interference, a specificity result that survives stratifying by numeric
token length, arguing against a generic quantization-noise account. A
trial-level error analysis further shows this specificity at the
mechanism level: quantization significantly raises the rate at which
wrong answers are same-key intrusions on an earlier, overwritten value,
rather than simply raising the overall error rate. A follow-up ablation
further localizes the effect within the model: quantizing the output
projection (lm\_head) on top of the backbone adds no detectable
effect beyond what backbone-only INT4 already produces, indicating the
penalty originates in the quantized transformer backbone rather than the
final vocabulary projection. The paired
re-analysis also revises the paper's original "cliff" narrative: INT4
carries a large, robust penalty in every model, while INT8 -- often
assumed safe -- carries a smaller but real penalty in two of the three
models once the appropriately-matched test is used. The overall effect is
invisible to aggregate benchmark accuracy, which differs by only 1--2
percentage points between FP16 and INT4 in our data, yet reaches 10--13
percentage points at the specific high-interference, semantically-confusable
conditions where it matters. All experiments were run on a single
consumer-grade GPU, which we hope makes this kind of behavioral,
non-aggregate evaluation more approachable for other researchers working
under similar hardware constraints. 

\bibliographystyle{unsrtnat}
\bibliography{references}

\appendix

\section{Full Task Design: Prompts, Scoring, and Vocabulary Construction}
\label{app:task-details}

\paragraph{Prompt template.}
Each trial consists of a sequence of natural-language update statements
followed by a retrieval question. Given an interference level $k$, exactly
$k$ update statements are generated for the same subject and the same
attribute, with each statement overwriting the previous value. The updates
appear in chronological order, and the final update always contains the
ground-truth value. The retrieval question is appended after a blank line
and explicitly instructs the model to report only the most recent value.
For example, a word-type trial takes the following form:

\begin{tcolorbox}[
  colback=gray!4,
  colframe=gray!55!black,
  boxrule=0.5pt,
  arc=2mm,
  left=8pt, right=8pt, top=6pt, bottom=6pt,
  title=Example prompt (word-type tria with $k$ equal to $3$),
  fonttitle=\bfseries\small,
  colbacktitle=gray!15,
  coltitle=black,
  fontupper=\ttfamily\small
]
Adam's mood is now happy.\\
Adam's mood is now nervous.\\
Adam's mood is now calm.

\vspace{6pt}
\normalfont\itshape Based only on the most recent update above, what is Adam's mood right now? Reply with a single word and nothing else.
\end{tcolorbox}

Numeric attributes use the same structure but replace the update template
and question accordingly (e.g., temperature or stock price) while
requesting only the final numeric value. Thus every prompt follows the
general pattern
\[
\underbrace{\text{Update}_1,\;
\text{Update}_2,\;
\ldots,\;
\text{Update}_k}_{\text{chronological overwrites}}
\;\rightarrow\;
\text{Retrieval Question}
\]
where the only correct answer is the value introduced in the final update,
and the prompt structure remains identical across all trials. Prompts are
provided to each model through its native instruction-chat template (e.g.,
Qwen, Mistral, and Phi chat formats) using the tokenizer's built-in
function, as a single user message with no system prompt.

\paragraph{Scoring and exact-match criteria.}
Models are prompted to answer with a single word (word-type attributes) or
a single number (numeric attributes) and nothing else, and generation is
capped at 12 new tokens. The raw response is lowercased and stripped of
leading/trailing whitespace and a single trailing period, if present. For
word-type attributes, we take the first contiguous alphabetic substring in
the cleaned response (via a simple regular expression) and compare it
against the gold value, also lowercased; a trial is scored correct only on
an exact string match, with no partial-credit or fuzzy matching. For
numeric attributes, we take the first contiguous run of digits (optionally
signed) in the cleaned response and compare it against the gold integer as
a string. Because generation is greedy and prompts explicitly instruct a
single-token-style answer, the overwhelming majority of responses consist
of exactly the target word or number with no additional content; the
regular-expression extraction step exists primarily to strip a stray
trailing period or an occasional short preamble rather than to resolve
ambiguous outputs. We apply no normalization beyond casing, whitespace, and
the trailing-period strip described above (e.g.\ no stemming, no synonym
matching, no numeric tolerance window), so scoring is deliberately
conservative in the same way across all three quantization levels.

\paragraph{Tokenizer-verified vocabulary construction.}
For word-type attributes, each interference level $k$ requires $k$ unique
values sampled without replacement from that attribute's vocabulary pool.
A naive approach risks two confounds as pools are made large enough to
support high $k$: (1) word rarity drift (later-added words being less
common), and (2) token-length drift (later-added words spanning more
subword tokens). Both would let interference level and item difficulty
covary, contaminating any accuracy-vs-level relationship. We mitigate this
by fixing one large candidate list per category (80--140 common, everyday
words) and filtering it, per tokenizer, to keep only words that occupy
exactly one token \emph{as they actually appear in the trial template}. We
determine this by tokenizing the fully instantiated template string and
checking, via the tokenizer's character-offset mapping, how many resulting
tokens overlap the candidate word's character span. This check is robust
across tokenizer families: byte-level BPE tokenizers (e.g.\ Qwen) fuse a
leading space into the following word's token, while SentencePiece-based
tokenizers (e.g.\ Mistral, Phi) may add a constant leading token
independent of content; both behaviors are correctly handled by reading
offsets directly off the real tokenized sentence rather than inferring
token count arithmetically from separately-encoded
fragments.\footnote{An earlier version of our pipeline used exactly such
an arithmetic approach (subtracting a "blank" baseline encoding) and
produced silently corrupted vocabulary pools for one tokenizer family
while appearing to work for another -- we flag this as a methodological
pitfall worth documenting for others building similar evaluations.} This
yields a fixed, per-model, per-category vocabulary pool whose maximum
usable interference level differs by model (Table \ref{tab:vocab-sizes})
but whose quality does not vary with $k$: every level for a given
model/category samples from the same fixed, filtered pool.

\begin{table}[h]
\centering
\caption{Maximum interference level tested per word-type attribute, by
model (limited by tokenizer-filtered single-token vocabulary size).}
\label{tab:vocab-sizes}
\begin{tabular}{lcccc}
\toprule
Model & mood & favorite color & favorite animal & occupation \\
\midrule
Qwen2.5-7B-Instruct        & 64 & 32 & 32 & 32 \\
Mistral-7B-Instruct-v0.3   & 32 & 16 & 16 & 16 \\
Phi-3.5-mini-instruct      & 16 & 16 & 8  & 16 \\
\bottomrule
\end{tabular}
\end{table}

Numeric attributes draw from a pool of $\sim$990 unique integers per
category and therefore support all tested interference levels for every
model. Unlike the word-type pools, we do not explicitly filter numeric
values to be single-token: integers in our range (3--998) can span one or
two tokens depending on the tokenizer and value. This does not reintroduce
the level-drift confound the word-type filtering was designed to prevent,
because the numeric pool is fixed and level-invariant -- every
interference level samples from the same range of integers, so token-count
composition does not systematically covary with $k$ the way an
insufficiently large word pool would. We nonetheless flag this as a minor
asymmetry in rigor between the two conditions (Section~\ref{sec:limitations}).

\section{Reproducibility: Model Loading Configuration}
\label{app:repro}

All experiments were conducted on a single NVIDIA RTX 3090 GPU with 24\,GB
of VRAM. Unless otherwise specified, all models were loaded using the
configuration summarized in Table~\ref{tab:model-loading-config}. No
additional dtype autocasting was applied beyond the compute dtype specified
for INT4 quantization.

\begin{table}[h]
\centering
\small
\caption{Model loading and quantization configuration used in all experiments.}
\label{tab:model-loading-config}
\begin{tabular}{ll}
\toprule
\textbf{Setting} & \textbf{Value} \\
\midrule
GPU & NVIDIA RTX 3090 (24\,GB VRAM) \\
Device map & \texttt{\{"":0\}} \\
Low CPU memory usage & \texttt{True} \\
\midrule
\multicolumn{2}{l}{\textbf{INT8 configuration}} \\
BitsAndBytes & \texttt{load\_in\_8bit=True} \\
LLM.int8 outlier threshold & 6.0 (default) \\
\texttt{llm\_int8\_skip\_modules} & None \\
\midrule
\multicolumn{2}{l}{\textbf{INT4 configuration}} \\
BitsAndBytes & \texttt{load\_in\_4bit=True} \\
Quantization type & NF4 \\
Double quantization & Enabled \\
Compute dtype & FP16 (\texttt{torch.float16}) \\
NF4 block size & 64 (default) \\
\texttt{modules\_to\_not\_convert} & None \\
\midrule
Additional dtype autocasting & None \\
\bottomrule
\end{tabular}
\end{table}

\section{Trial Budget, Seeds, and Paired Design}
\label{app:trials}

Each experimental condition (model $\times$ quantization $\times$
interference level $\times$ attribute) was evaluated using five
independent random seeds, resulting in a total of 69{,}225 trials
(28{,}050 word-type and 41{,}175 numeric) across 45
(model $\times$ quantization $\times$ seed) runs.

For a fixed (model, seed) pair, the complete sequence of generated trials
(including subject names, distractor values, trial ordering, attributes,
and interference levels) was identical across FP16, INT8, and INT4. Thus,
each quantization level was evaluated on exactly the same set of prompts,
with the model weights being the only difference between runs. Consequently,
all FP16--INT8 and FP16--INT4 comparisons reported in this paper are
performed using paired trials rather than independently sampled trials,
eliminating variation in item difficulty as a potential confounding factor.

Across different seeds, subject names and distractor values were sampled
independently from the same fixed, tokenizer-filtered vocabulary pool.
Finally, the trial budget was intentionally weighted toward higher
interference levels, where floor and ceiling effects are less likely to
obscure the impact of quantization (Table~\ref{tab:trial-budget}).

\begin{table}[h]
\centering
\caption{Trials per interference level, per (model, quantization, seed, attribute) cell.}
\label{tab:trial-budget}
\begin{tabular}{lcccccccc}
\toprule
Level $k$ & 1 & 2 & 4 & 8 & 16 & 32 & 64 & 96 \\
\midrule
Trials & 15 & 15 & 20 & 25 & 60 & 50 & 60 & 60 \\
\bottomrule
\end{tabular}
\end{table}

\section{Statistical Methodology: Full Details}
\label{app:stats-detail}

For each model, we identify the interference level with the greatest
statistical power for detecting a quantization effect -- i.e.\ farthest
from ceiling (ceiling effects dominate at low $k$ for all models) and
farthest from floor (floor effects dominate at very high $k$ for the
weaker models): Qwen at $k=64$, Mistral at $k=16$, and Phi at $k=8$. At
each selected level we compare FP16 vs.\ INT4 accuracy via Fisher's exact
test and report the odds ratio with a 95\% confidence interval (computed
with statsmodels; \citealp{seabold2010statsmodels}), and separately
confirm the direction of the effect (FP16 $>$ INT4) is consistent across
all 5 seeds individually. To test whether the effect generalizes across
models rather than relying on three separate per-model tests, we compute a
Mantel--Haenszel test \citep{mantelhaenszel1959}, pooling the three
models' $2\times2$ tables into a single common odds ratio, confidence
interval, and significance test that controls for model identity as a
stratifying variable.

Second, and more importantly, because the per-model level selection above
is chosen post-hoc for statistical power (a reasonable but auditable
choice that a skeptical reader might still flag as researcher degrees of
freedom), we additionally fit a mixed-effects logistic regression using
\emph{all} interference levels simultaneously, for every model and seed at
once:
\[
\text{correct} \sim \text{quant\_mode} * \log_2(\text{interference\_level}) + (1\mid\text{model}) + (1\mid\text{seed}) + (1\mid\text{attribute})
\]
fit via variational Bayes using a Bayesian mixed generalized linear model
(statsmodels; \citealp{seabold2010statsmodels}), with FP16 as the
reference level for quantization mode and random intercepts for model,
seed, and attribute. The attribute-level random intercept is included
because our word-type and numeric attributes differ substantially in
baseline difficulty (e.g.\ mood is markedly harder than favorite animal
at the same interference level); without it, this attribute-level
heterogeneity could in principle be partly absorbed into the fixed
effects and inflate their apparent significance. The key term is the
quantization-mode $\times$ log-interference-level interaction: a
significant negative coefficient for INT4 means the INT4 accuracy penalty
relative to FP16 grows as interference increases, using the complete
dataset rather than any selected subset. We fit this model separately for
word-type and numeric attributes, since the specificity claim predicts
opposite signs for the two.

\paragraph{Paired significance testing (McNemar's test).} The per-model
comparisons above (Fisher's exact test, Mantel--Haenszel) are unpaired:
they compare aggregate correct/incorrect counts between two quantization
levels without exploiting the fact that, by construction, the same trial
content is presented to both. We verified this pairing holds exactly
across our full dataset: for every one of the 23{,}075 unique (model,
seed, attribute, level, trial) combinations, the logged subject name and
gold value are identical across all three quantization levels, with zero
mismatches. Given this, we report McNemar's exact test as our primary
per-model significance test at each model's selected level: this test uses
only the discordant pairs (trials on which the two compared quantization
levels disagree) and is therefore substantially better powered than the
marginal Fisher's exact test, which discards the pairing entirely. We
retain the unpaired Fisher's-exact/Mantel--Haenszel results as a more
conservative cross-check that does not depend on the pairing assumption.

\paragraph{Random-slopes robustness check.} To rule out the possibility
that a model-specific or attribute-specific interference-level slope is
being absorbed into the fixed-effect interaction terms of the mixed model
above, we refit it with two additional variance components: random slopes
for $\log_2(\text{level})$ by model and by attribute, alongside the random
intercepts already described.

\paragraph{Error-type classification (same-key intrusion analysis).}
Because trial generation is fully deterministic given its random seed, we
reconstruct the exact ordered sequence of overwritten distractor values
for every trial by replaying the seeded sampling procedure, and confirm
this reconstruction exactly reproduces the logged subject name and gold
value for all 23{,}075 unique trials, with zero mismatches. This lets us
classify every incorrect response as a \emph{same-key intrusion} (the
extracted answer exactly matches one of the $k-1$ overwritten values for
that trial) or a non-matching error, and to compute an intrusion rate as a
fraction of all trials, not just of errors.

\paragraph{Numeric token-length stratification.} To address the
token-length asymmetry between our tokenizer-filtered word-type pools and
the unfiltered numeric pool, we separately tokenize every numeric value
actually used in a trial, in its real template context, per model
tokenizer, using the same offset-mapping method as the word-type
filtering. We use this to check whether token-length composition drifts
with interference level (which would reintroduce the confound the
word-type filtering was designed to avoid) and to stratify the
FP16-vs-INT4 numeric comparison by token length.

\paragraph{Interference Endurance Score (IES).} Following \citet{wang2025unable},
we define IES as the area under the accuracy-vs-interference-level curve,
computed via trapezoidal integration over a log$_2$-scaled interference
axis. We report a normalized variant -- raw AUC divided by the tested
log$_2$ range -- so that IES falls on the same $[0,1]$ scale as accuracy
and remains comparable across models despite differing maximum tested
interference levels (Table~\ref{tab:vocab-sizes}). IES is computed
directly from the paper's existing trial-level dataset (Section~\ref{sec:stats}),
requiring no new experiments.

\section{Additional Results and Robustness Checks}
\label{app:additional-results}

\paragraph{Phi non-monotonicity at $k=16$.} Table \ref{tab:phi-anomaly}
shows the one condition in our data where INT8 accuracy falls below INT4.
We report this rather than omit it; it does not change the paper's overall
conclusions, since Phi's key-level comparison (Table \ref{tab:key-results},
Table \ref{tab:int8-comparison}) is conducted at $k=8$, not $k=16$.

\begin{table}[h]
\centering
\caption{Phi-3.5-mini-instruct at $k=16$: the one condition in our data
where INT8 accuracy falls below INT4, $n=300$ per cell.}
\label{tab:phi-anomaly}
\begin{tabular}{lccc}
\toprule
 & FP16 & INT8 & INT4 \\
\midrule
Accuracy & 37.1\% & 30.7\% & 35.3\% \\
\bottomrule
\end{tabular}
\end{table}

\paragraph{Interpreting the mixed-model main effects.} Beyond the
interaction terms reported in Table \ref{tab:mixed-model}, the model also
shows sizeable negative main effects for both INT8 and INT4 relative to
FP16, in both word-type and numeric attributes. Because interference
level is log-transformed and centered at $\log_2(1)=0$, these main-effect
coefficients describe the quantization gap specifically at the lowest
interference level ($k=1$), where accuracy is near ceiling for every
model and condition (Table \ref{tab:overall-accuracy}, Figure
\ref{fig:accuracy-by-level}). A significant main effect alongside a
near-ceiling raw accuracy simply reflects that logistic regression
coefficients operate on the log-odds scale, where small absolute
probability differences near 100\% correspond to comparatively large
log-odds differences; it should not be read as evidence of a practically
meaningful gap at $k=1$, where raw accuracy differences between
quantization levels are well under one percentage point in our data. The
interaction terms, not the main effects, are the theoretically relevant
quantities for this paper's hypothesis, since they capture how the
quantization gap \emph{changes} with interference rather than its value
at one arbitrary reference level.

\paragraph{Robustness to random slopes.} Table \ref{tab:mixed-model}
includes random intercepts for model, seed, and attribute, but not random
\emph{slopes} for interference by model or attribute -- leaving open the
possibility that a model-specific or attribute-specific interference trend
is being absorbed into the fixed-effect interaction terms. Refitting with
random slopes for $\log_2(\text{level})$ by both model and attribute
leaves the key interaction essentially unchanged in both panels: for
word-type, the INT4 $\times$ log-interference-level interaction becomes
$\beta=-0.048$, OR $=0.95$, 95\% CI $[0.94, 0.97]$, $z=-5.62$,
$p=1.9\times10^{-8}$ (vs.\ $\beta=-0.046$, $p=5.7\times10^{-8}$ without
random slopes); for numeric, it becomes $\beta=+0.059$, OR $=1.06$, 95\%
CI $[1.05, 1.07]$, $z=11.84$, $p<10^{-16}$ (vs.\ $\beta=+0.053$,
$p<10^{-16}$). The opposite-signed specificity result -- this paper's
central claim -- is therefore not an artifact of unmodeled model- or
attribute-specific interference trends. One secondary term does change:
the numeric INT8 $\times$ log-interference-level interaction, weakly
significant without random slopes ($p=0.0049$), becomes non-significant
with them ($\beta=+0.004$, $p=0.369$); this term is not part of the main
hypothesis (which concerns INT4), so it does not affect our conclusions.

\paragraph{Ruling out a token-length confound.} Because the numeric pool
is not tokenizer-filtered to be single-token, unlike the word-type pools,
a skeptical reading of the specificity result is that it reflects
token-length differences rather than semantic (non-)confusability. Two
checks address this. First, the token-length composition of the numeric
values actually used is essentially flat across all eight interference
levels (dominated by 3-token numbers at 89.9--92.4\% of trials at every
level, with 2-token and 1-token values each maintaining a roughly constant
share), confirming the numeric pool does not exhibit the level-drift
confound the word-type filtering was designed to prevent. Second,
stratifying the FP16-vs-INT4 comparison at high interference levels (32,
64, 96 pooled) by token length shows the specificity direction holds in
every stratum and is individually significant in the dominant 3-token
stratum, which accounts for $>$89\% of all numeric trials: FP16 69.4\% vs.\
INT4 71.2\% ($n=6{,}882$ pairs, OR $=0.92$, 95\% CI $[0.85, 0.99]$,
$p=0.021$). The 1-token ($n=32$) and 2-token ($n=736$) strata point in the
same direction but are individually underpowered (OR $=0.86$, $p=0.78$;
OR $=0.94$, $p=0.57$). The numeric-attribute asymmetry flagged in
Section~\ref{sec:limitations} does not appear to drive the specificity
result.

\paragraph{IES on numeric (control) attributes.} Table
\ref{tab:ies-numeric} reports the same INT8/INT4 IES penalty computed on
numeric attributes, for comparison with the word-type results in
Section~\ref{sec:ies}. The pattern only partially corroborates the
specificity result of Section~\ref{sec:specificity}. For Mistral, the
INT4 penalty is clearly larger under word-type interference (5.1\%) than
numeric (2.6\%), consistent with the paper's central claim. For Qwen,
both penalties are small and comparable (1.8\% vs.\ 0.05\%), which is
consistent with the cliff-dilution effect discussed above rather than
with a genuine word/numeric asymmetry at this model's tested levels. For
Phi, INT4's numeric IES is \emph{higher} than FP16's ($-3.98\%$, i.e.\
INT4 nominally outperforms FP16 on the numeric control by this metric);
this is a small-magnitude reversal consistent with the sign flip already
established for numeric attributes in the mixed-effects analysis
(Section~\ref{sec:mixed-model}), and we treat it as a null/mixed result
for this model rather than evidence against the paper's specificity
claim. Phi's INT8 numeric penalty (5.49\%) is notably larger than its
word-type INT8 penalty (3.53\%); we have not independently traced this to
a specific interference level and report it as an observed, unexplained
asymmetry rather than folding it into the main narrative.

\begin{table}[h]
\centering
\caption{INT8/INT4 IES penalty (\% lost relative to FP16) on numeric
(control) attributes, for comparison with the word-type results in
Table~\ref{tab:ies-results}.}
\label{tab:ies-numeric}
\begin{tabular}{lcc}
\toprule
Model & INT8 \% IES lost & INT4 \% IES lost \\
\midrule
Qwen2.5-7B-Instruct      & $-0.07\%$ & $0.05\%$ \\
Mistral-7B-Instruct-v0.3 & $0.76\%$  & $2.60\%$ \\
Phi-3.5-mini-instruct    & $5.49\%$  & $-3.98\%$ \\
\bottomrule
\end{tabular}
\end{table}

\paragraph{A methodological detail not independently re-verified.} IES
for word-type attributes pools across four attributes (mood, favorite
color, favorite animal, occupation) whose tokenizer-filtered vocabulary
caps differ within a model (Table~\ref{tab:vocab-sizes}: e.g.\ for Qwen,
mood reaches $k=64$ but occupation caps at $k=32$). We did not verify how
the underlying computation handles this asymmetry -- whether the highest
tested levels are pooled only across the attributes that support them, or
some other convention -- and report the IES values in
Section~\ref{sec:ies} as provided, flagged here as an unverified detail
rather than a confirmed part of the methodology.

\section{lm\_head Ablation: Additional Methodological Detail}
\label{app:lmhead-narrative}

Our first attempt at this ablation reran the INT4 configuration with the
output projection (lm\_head) explicitly excluded from quantization via a
skip-list argument. This produced outputs identical to the original INT4
condition on all 5{,}050 paired trials, token for token. Inspecting the
loaded model revealed why: under the installed library version, lm\_head is
left at full precision by default irrespective of this argument. This means
the INT4 condition used throughout the paper never quantized lm\_head in
the first place -- a property of the library's defaults, not an error in
our setup -- so every INT4 result in the main text reflects backbone-only
quantization.

To obtain the intended comparison, we instead manually converted lm\_head
to a genuine 4-bit layer on top of the standard INT4 model, matching the
backbone's quantization settings, and confirmed the conversion took effect
before running any trials. This corrected condition differs from the
original backbone-only INT4 condition on 4.0\% of paired trials
(203/5{,}050) -- a small but genuine footprint, as expected from quantizing
one additional layer, and the basis for the comparison reported in
Section~\ref{sec:lmhead-ablation}. Full implementation details are
available in the released code repository.

\end{document}